\documentclass[runningheads]{llncs}

\usepackage{eccv}

\usepackage{eccvabbrv}

\usepackage{graphicx}
\usepackage{booktabs}

\usepackage[accsupp]{axessibility}  

\usepackage{amsmath}

\usepackage{tabularx}
\usepackage{tabularray}
\usepackage{wrapfig, lipsum, booktabs}
\usepackage{multirow}
\usepackage{float}

\usepackage{overpic}
\usepackage{bm}
\usepackage{mathtools}
\usepackage{mathtools, cuted}

\usepackage{makecell}

\usepackage{hyperref}

\usepackage{orcidlink}

\usepackage[normalem]{ulem}
\usepackage{tikz}
\usepackage{algorithm}
\usepackage{algpseudocode}
\newcommand{\DashedLine}{%
\Statex\hspace*{-\algorithmicindent}\tikz{\draw[dashed] (0,0) -- (1.07\linewidth,0);}%
}

\begin{document}

\title{Surf-D: Generating High-Quality Surfaces of Arbitrary Topologies Using Diffusion Models} 

\titlerunning{Surf-D}

\author{Zhengming Yu\inst{1}{\thanks{Equal Contribution}}\orcidlink{0009-0003-0553-8125} \and Zhiyang Dou\inst{2,6}$^{\star}$\orcidlink{0000-0003-0186-8269} \and Xiaoxiao Long\inst{2, 3}\orcidlink{0000-0002-3386-8805} \and Cheng Lin\inst{2}\orcidlink{0000-0002-3335-6623
} \and \\Zekun Li\inst{4}\orcidlink{0000-0002-1461-4121} \and Yuan Liu\inst{2}\orcidlink{0000-0003-2933-5667} \and Norman M\"uller\inst{5}\orcidlink{0009-0006-8034-9721} \and Taku Komura\inst{2}\orcidlink{0000-0002-2729-5860} \and \\Marc Habermann\inst{3}\orcidlink{0000-0003-3899-7515} \and Christian Theobalt\inst{3}\orcidlink{0000-0001-6104-6625} \and Xin Li\inst{1}\orcidlink{0000-0002-0144-9489} \and Wenping Wang\inst{1}\orcidlink{0000-0002-2284-3952}}

\authorrunning{Z. Yu, Z. Dou et al.}

\institute{Texas A\&M University \and
The University of Hong Kong \and
Max Planck Institute for Informatics \and
Brown University \and Meta Reality Labs Zurich \and TransGP}

\maketitle

\begin{abstract}
We present \textbf{Surf-D}, a novel method for generating high-quality 3D shapes as \textbf{Surfaces} with arbitrary topologies using \textbf{Diffusion} models. 
Previous methods explored shape generation with different representations and they suffer from limited topologies and poor geometry details. To generate high-quality surfaces of arbitrary topologies, 
we use the Unsigned Distance Field (UDF) as our surface representation to accommodate arbitrary topologies. Furthermore, we propose a new pipeline that employs a point-based AutoEncoder to learn a compact and continuous latent space for accurately encoding UDF and support high-resolution mesh extraction. We further show that our new pipeline significantly outperforms the prior approaches to learning the distance fields, such as the grid-based AutoEncoder, which is not scalable and incapable of learning accurate UDF. In addition, 
we adopt a curriculum learning strategy to efficiently embed various surfaces. With the pretrained shape latent space, we employ a latent diffusion model to acquire the distribution of various shapes. Extensive experiments are presented on using Surf-D for unconditional generation, category conditional generation, image conditional generation, and text-to-shape tasks. The experiments demonstrate the superior performance of Surf-D in shape generation across multiple modalities as conditions. Visit our project page at \url{https://yzmblog.github.io/projects/SurfD/}.
\end{abstract}

\section{Introduction}
\label{sec:intro}

\begin{figure}[tb]
\centering
\begin{overpic}[width=1.05\linewidth]{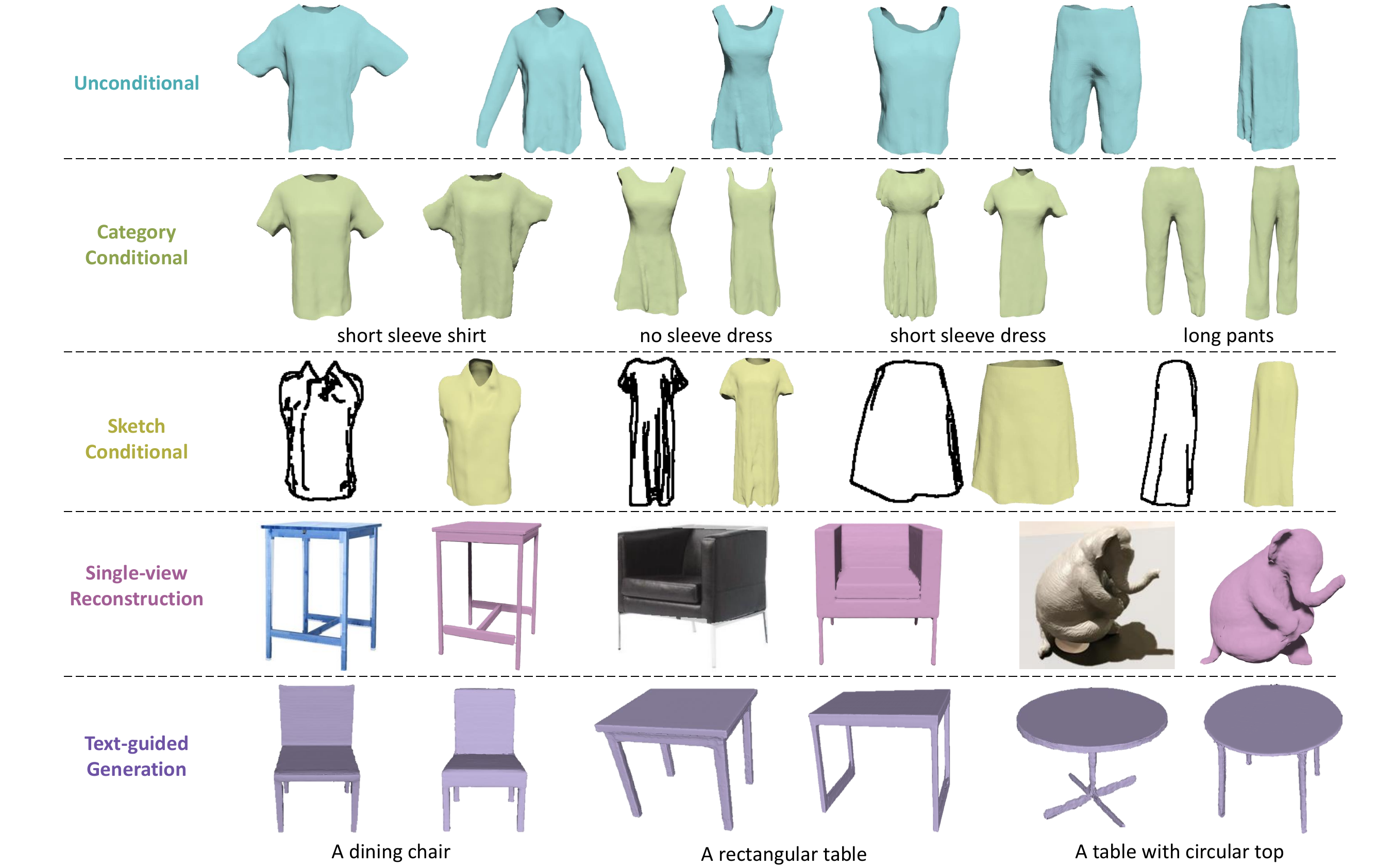}
\end{overpic}
    \caption{\textbf{Surf-D} achieves high-quality \textbf{Surface} generation for detailed geometry and various topology using a \textbf{Diffusion} model. It achieves SOTA performance in various shape generation tasks, including unconditional generation, category conditional generation, sketch conditional shape generation, single-view reconstruction, and text-guided shape generation.}
\label{fig:teaser}
\end{figure}

Generating a diverse array of 3D shapes using various conditions is an important problem in computer vision and graphics, given its wide applications, such as gaming, filmmaking, and the metaverse. 

\begin{wrapfigure}[16]{r}{5.5cm}
  \begin{center}
  \begin{overpic}[width=0.4\textwidth]{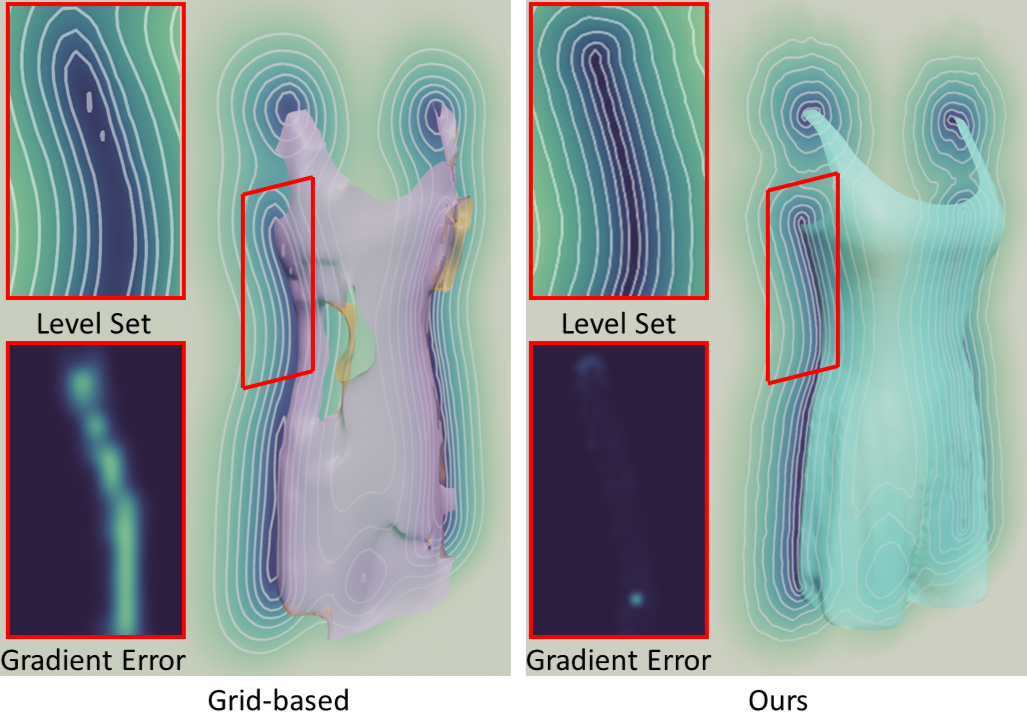}
\put(24,-6.5){(a)} 
\put(72,-6.5){(b)}
\end{overpic}
  \end{center}
  \caption{Comparison of learning discrete grid-based field~\cite{cheng2023sdfusion} and continuous field (ours) for UDFs (top) and corresponding gradient errors (bottom).The darker the color, the smaller the values.}
  \label{fig:1}
\end{wrapfigure}

Recent advancements for 3D shape generation \cite{chan2021pi, achlioptas2018learning, luo2021diffusion, li2023neuralangelo, gu2021stylenerf, liu2023syncdreamer, long2023wonder3d, liu2024part123} aim to make this process more accessible. The diverse interplay of topology and intricate geometry gives rise to countless 3D shapes in our world.  However, existing generation works still suffer from limited topologies and the lack of geometric details. Previous methods present shapes using 
point clouds \cite{achlioptas2018learning, luo2021diffusion}, voxel representations \cite{smith2017improved, xie2018learning}, and signed distance fields (SDFs) \cite{chen2019learning, autosdf2022}. However, point cloud and voxel representations encounter difficulties in faithfully extracting surfaces. Implicit representations, particularly SDFs, excel in describing closed surfaces but face challenges with open surfaces and complex topologies. Despite attempts to enhance them, \eg 3PSDF~\cite{chen20223psdf} and TSDF~\cite{sun2021neuralrecon}, the representations remain reliant on the underlying definition of SDF. G-Shell~\cite{liu2023ghost} is proposed to represent open surfaces but still falters with non-manifold structures and computational constraints.

Previous 3D shape generation methods based on learning a distance field include SDFusion~\cite{cheng2023sdfusion} and its variants~\cite{ li2023diffusion, zheng2023lasdiffusion}. These methods consider only SDF and all use a discrete 3D feature grid to approximate the SDF field.  

This grid-based approach is fundamentally limited and unsuitable for learning high-quality UDF because of the following drawbacks: (\romannumeral1)~it does not provide access to accurate and stable gradient of the distance field, which is essential for accurately reconstructing the surface encoded by the UDF; (\romannumeral2)~limited resolution of the generated grid;  (\romannumeral3)~expensive memory cost for the grid representation.

With these disadvantages, the grid-based framework cannot be used for learning UDF to generate high-quality surfaces of arbitrary topologies, as confirmed by our experiments. 
Briefly, Fig.~\ref{fig:1} shows that the UDF generated using the grid-based method~\cite{cheng2023sdfusion} suffers from unstable, as shown in the gradient field error map\footnote[1]{We calculate the numerical gradients of UDF following Neuralangelo~\cite{li2023neuralangelo} for visualizing the error map and meshing.} in Fig.~\ref{fig:1} (a) bottom, in comparison with our results in Fig.~\ref{fig:1} (b) bottom. These unreliable gradients will in turn lead to low-quality surface meshes.

To tackle the aforementioned problems, We propose \textbf{Surf-D}, the first generative latent diffusion model~\cite{stable-diffusion} employing Unsigned Distance Fields (UDF) which is a general shape representation for any 3D shape, for versatile shape generation with high-quality surfaces of arbitrary topologies. Unlike previous discrete grid-based shape embedding methods~\cite{cheng2023sdfusion,li2023diffusion,zheng2023lasdiffusion}, Surf-D utilizes a point-based AutoEncoder to effectively capture UDFs by compressing the underlying 3D surfaces into compact low-dimensional latent codes
and learns \textit{Continuous Distance Fields}, which allows for a more accurate and robust gradient field computation, as evidenced by the gradient error map in Fig.~\ref{fig:1} (b) bottom. 
Compared with existing discrete-grid-based approaches, our Surf-D enjoys several advantages: (\romannumeral1)~reliable gradient query for each input point which supports accurate mesh extraction~\cite{guillard2022meshudf}; (\romannumeral2)~capturing UDFs without a restriction by resolution, as it learns a continuous field; 
 (\romannumeral3)~high computational efficiency; See detailed experiments and analysis between our approach and discrete grid-based shape embedding in Sec.~\ref{exp:ablation_autoencoder}.

Moreover, to learn the compact shape latent space efficiently and effectively on diverse surfaces, we employ a curriculum learning strategy~\cite{bengio2009curriculum} to train our model in an easy-to-hard sampling order, \ie progressively extending the current training dataset during training. This strategy promotes our model to develop a diverse latent space, avoiding entrapment into generating mean shape, \ie model collapse. As a result, our model is not only lightweight and learning-efficient but also feasible to convert high-quality generated implicit shapes to explicit meshes.

We conduct extensive experiments across wide datasets, where our model outperforms prior work quantitatively and qualitatively in shape generation, 3D reconstruction from images, and text-to-shape tasks, as shown in Fig.~\ref{fig:teaser}.

The main contributions of this paper are as follows:
\begin{enumerate}
    \item We propose Surf-D, the first surface generation model based on the diffusion model using UDFs to generate 3D shapes with arbitrary topology and detailed geometry.
    \item We conduct latent diffusion and devise a point-based AutoEncoder that enables effective gradient querying for given input points in the field, achieving high-resolution and detail-preserving surface generation. To enhance surface learning, we implement a curriculum learning scheme.
    \item  Surf-D achieves superior performance in generating 3D shapes with diverse topologies while enabling various modalities conditional generation tasks.
\end{enumerate}

\section{Related Work}
\subsection{3D Shape Generation}

Tremendous efforts have been made for 3D shape generation~\cite{chan2021pi, achlioptas2018learning, luo2021diffusion, chan2022efficient, gu2021stylenerf,liu2023meshdiffusion,chen2023single, liu2024part123, li2024era3d, li2024craftsman, yang2022dsg, gao2019sdm, wang2019learning, sun2023dreamcraft3d}, given its broad applications, employing various shape representations such as voxel grids~\cite{wu20153d,wu2016learning}, point clouds~\cite{luo2021diffusion, vahdat2022lion, zhou20213d}, and implicit neural fields~\cite{park2019deepsdf,duan2020curriculum,cheng2023sdfusion,hui2022neural, long2023neuraludf, meng2023neat, liu2023neudf, wang2022hsdf, chen20223psdf, tang2021octfield, shen2023flexible, shen2021deep}. Additionally, various methods have been developed to guide the synthesis with text prompt~\cite{sanghi2023clip,fu2022shapecrafter,liu2023exim,Chen_2023_ICCV,liu2023dreamstone, wang2023disentangled}, single-view image~\cite{liu2023syncdreamer,long2023wonder3d,liu2023zero,liu2023one2345,wu2023star}, sketch~\cite{wu2023sketch,zheng2023lasdiffusion} and sparse control anchors~\cite{lyu2023controllable,Chou_2023_ICCV}. In the early stages, voxel-based generation methods~\cite{wu20153d,wu2016learning,vpp2023} can synthesize various coarse shapes. However, the voxel representation can not scale to high-resolution generation~\cite{zheng2022neural,zheng2023lasdiffusion}, due to its cubic space complexity.
Utilizing point clouds~\cite{luo2021diffusion, vahdat2022lion, zhou20213d}, which are compact and readily available, for 3D object representation can achieve arbitrary shape generation with their well-suitability to learn geometry and texture information~\cite{nichol2022point,wu2023sketch}. However, it requires complex post-processing to faithfully convert the synthesized point clouds into meshes for real-world applications~\cite{peng2021shape}. The implicit representation such as SDF~\cite{cheng2023sdfusion} and differentiable explicit shape representations such as FlexiCubes~\cite{shen2023flexible} and DMTet~\cite{shen2021deep} can achieve detailed and high-quality results. However, they perform poorly in modeling the non-watertight shapes as their definitions are based on SDF. Also, they are highly memory-consuming when incorporated with diffusion models.

\subsection{Diffusion Probabilistic Models}
Diffusion probabilistic models~\cite{ho2020denoising} have demonstrated superior performance in generating samples from a distribution by learning to denoise from a data point gradually~\cite{dhariwal2021diffusion,dalle2,voleti2022mcvd,tevet2022human}. The diffusion models produce high-quality contents including images~\cite{liu2023syncdreamer, dhariwal2021diffusion, improve_ddpm, dalle2, stable-diffusion}, video~\cite{voleti2022mcvd, he2022latent, ge2023preserve}, and human motion~\cite{tevet2022human, karunratanakul2023gmd,shi2023controllable, zhou2023emdm}.
Previous shape generation works~\cite{liu2023meshdiffusion,hui2022neural} directly train diffusion models on massive grids storing shape representation like SDF, consuming plenty of GPUs and taking days for training. To improve the efficiency, the latent diffusion approach~\cite{stable-diffusion} has been explored in 3D shape generation. However, naively applying the latent diffusion model leads to poor generation capability as the vanilla neural field is easy to overfit a single shape. Thus, many efficient generation frameworks are proposed to promote the learning of generative neural networks, \eg coarse-to-fine generation~\cite{zheng2023lasdiffusion,shim2023diffusion,vahdat2022lion}, feature grids with codebooks~\cite{cheng2023sdfusion,fu2022shapecrafter,Li_2023_CVPR}, tri-planes encoding~\cite{wu2023sin3dm,Chou_2023_ICCV}, and primitive embedding~\cite{Koo_2023_ICCV}. In this paper, similar to \cite{zhao2023michelangelo}, we first pre-train a shape point-based AutoEncoder to learn the high-fidelity shape embedding of UDFs, providing a compact latent space for our latent diffusion model to synthesize various realistic shapes.
\begin{figure*}[!t]
    \centering
\includegraphics[width=\linewidth]{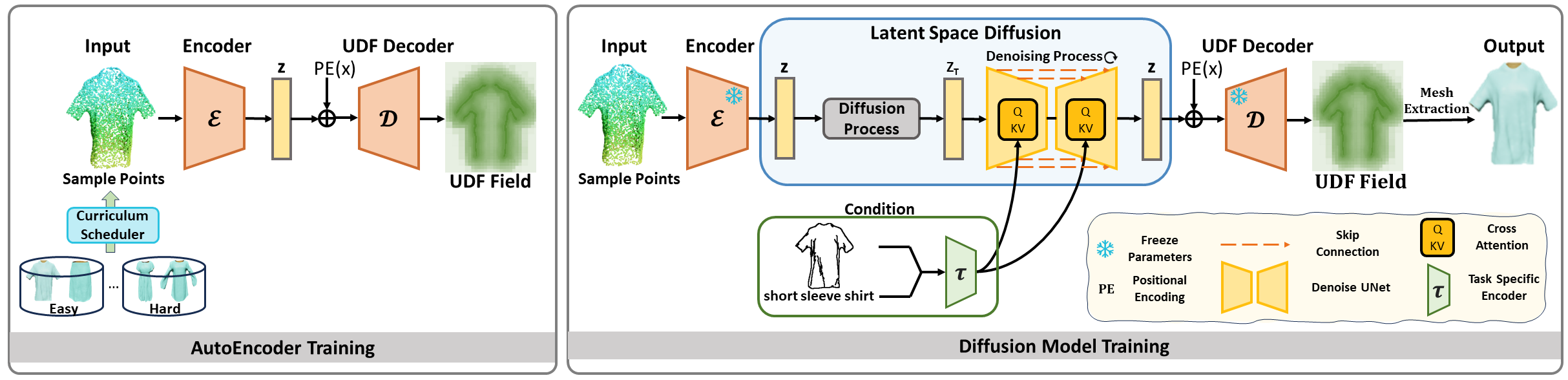}
\caption{\textbf{Framework of Surf-D.} We first encode surface points into a latent code $z$ by our encoder $\mathcal{E}$. A curriculum scheduler helps to train our model in an easy-to-hard sample order. Then we train the diffusion model in our latent space and various conditions can be added by a task-specific encoder $\tau$. Finally, the sampled latent code $z$ will be decoded to a UDF field for mesh extraction by our UDF decoder $\mathcal{D}$.}
    \label{fig:pipeline}
\end{figure*}

\section{Method}
\subsection{Point-based AutoEncoder}
\label{sec:shape_embed}
\paragraph{Network Structure}
\label{sec:network_structure}

To effectively learn a compact latent space for UDFs, we use a point cloud encoder to
learn a continuous field of UDFs, instead of directly taking the distance value grid as input like previous works~\cite{cheng2023sdfusion, li2023diffusion, zheng2023lasdiffusion}. The encoder embeds points sampled from the 3D object surface into a condensed vector, which can be done given any arbitrary typology surface. Then, the encoder is coupled with a decoder that accepts the latent vector and the query coordinate as inputs and outputs the unsigned distance between the query point and the encoded shape. With this structure, the information of sampled point clouds, together with the latent code, could effectively capture the topology and geometric details of the surface. This allows us to obtain more reliable pseudo-signs in bisection search~\cite{guillard2022meshudf} for surface extraction from UDFs, which thus significantly enhances the overall quality of generated surfaces in terms of both topology and geometry.

Specifically, we begin by sampling $N$ points $\mathcal{P} \in \mathbb{R}^{N\times 3}$ on the 3D surface. We utilize a point cloud encoder~\cite{wang2019dynamic} for feature extraction, leveraging its ability to propagate features within the same local region at multiple scales and aggregate them into a unified global embedding. The compressed latent codes then pass to our points query decoder, a Conditional Batch Normalization-based multi-layer perceptron (MLP)~\cite{de2017modulating}, for decoding UDF fields conditioned on the input latent vector. Mathematically, the input point set $\mathcal{P}$ is encoded by the encoder $\mathcal{E}$ into a latent space $\mathbf{z} \in \mathbf{R}^{d}$. The latent code $\mathbf{z}$ is subsequently passed to the decoder $\mathcal{D}$ along with the positional encoding~\cite{mildenhall2020nerf} of 3D point query $x$ to predict the UDF value $\widetilde{y}$. 
\begin{equation}
    \widetilde{y} = \mathcal{D}(\mathcal{E}( \mathcal{P}), PE(x))
    \label{eq:vavae}
\end{equation}

\paragraph{Loss Functions}
\label{sec:loss_functions}
Next, we detail the loss functions used for point-based AutoEncoder training. We first normalize the UDF field into $[0,1]$ distribution and employ binary cross-entropy as our reconstruction loss. Specifically, given a set of query points $\mathcal{X}=\{\mathbf{x}_{i} | \mathbf{x}_{i}\in \mathbb{R}^3\}$ sampled from the space surrounding the object, we clip all the ground-truth distance values $\{y_{i}\}$ to a distance threshold $\delta$, and linearly normalize the clipped values to the range $[0,1]$. This yields normalized ground-truth values $\bar{y}_{i}$ $=\min(y_{i}, \delta)/\delta$. Following~\cite{duan2020curriculum, de2023drapenet}, we set $\delta=0.1$. Formally, the reconstruction loss can be represented as follows:

\begin{equation}
    \mathcal{L}_{recon} =  \mathbb{E} \left[ -{(\bar{y}_{i}\log(\widetilde{y}_{i}) + (1 - \bar{y}_{i})\log(1 - \widetilde{y}_{i}))} \right] \;.
    \label{eq:gen_dist_loss}
\end{equation}

Additionally, our model adopts gradient supervision to achieve better geometry details of implicit shape learning. Given the same sample points as before, we take the gradient loss to be
\begin{equation}
    \mathcal{L}_{grad} = \mathbb{E} \left[ \left \| \mathbf{g}_{i} - \widehat{\mathbf{g}_{i}} \right \|_2^2\right] \;,
    \label{eq:gen_grad_loss}
\end{equation}
where $\mathbf{g}_{i} = \nabla_{\mathbf{x}} y_{i} \in \mathbb{R}^3$ is the ground-truth gradient of the shape's UDF at $\mathbf{x}_{i}$ and $\widehat{\mathbf{g}_{i}} = \nabla_{\mathbf{x}} \mathcal{D}(\mathbf{x}_{i}, \mathbf{z})$ is the one of the predicted UDF, computed by backpropagation. The total loss is given by
\begin{equation}
  \mathcal{L}_{latent}  =   \mathcal{L}_{recon}  + \lambda  \mathcal{L}_{grad} \;.
    \label{eq:loss_latent}
\end{equation}

With the proposed point-based AutoEncoder, our shape embedding captures a distribution of various shapes represented by the UDFs effectively, allowing for efficient shape sampling within the diffusion model in the following generation tasks. The powerful capability of the latent space we learned is shown in Sec.~\ref{exp:latent_interpolation}. We also investigate different network structures in Sec.~\ref{exp:ablation_autoencoder} to demonstrate the superiority of the point-based AutoEncoder.

\paragraph{Curriculum Learning Strategy}
\label{sec:curriculum}
To enhance the learning performance on diverse shapes with arbitrary topologies, we implement curriculum learning instead of training directly on the entire shape dataset holistically. Curriculum learning involves training machine learning models in a meaningful order, from easy to challenging samples, known as hard sample mining. This approach can enhance performance without incurring additional computational costs~\cite{bengio2009curriculum, soviany2022curriculum}.

Curriculum training involves measuring performance and establishing an easy-to-hard scheduler. In this paper,  we use a simple yet effective performance metric, which is the training loss $\mathcal{L}_{latent}$. We select samples with lower losses as incoming training data in each iteration. This continues until all data is included. Specifically, we first randomly sample $k$ shapes from the dataset, then gradually add new difficult samples according to their loss values, resulting in an easy-to-hard gradual learning process. The pseudo-code can be expressed as Algo.~\ref{algo:curriculum_alg}. The influence of the proposed learning strategy is included in the Appendix.
\begin{figure}
    \centering
    \begin{minipage}{\textwidth}
\begin{algorithm}[H]
\caption{Curriculum Learning Strategy for Shape Embedding}
\label{algo:curriculum_alg}
\begin{algorithmic}[1]
 \renewcommand{\algorithmicrequire}{\textbf{}}
 \Statex $M$ -- our model;
 \Statex $\mathcal{D}$ -- the whole dataset;
 \Statex $\mathcal{D}^*$ -- current training dataset;
 \Statex $n$,$s$,$k$ -- the number of training epochs, switching epochs, and selected samples;
 \Statex $\mathcal{L}$ -- Loss calculation;
 \Statex $\mathop{\mathrm{arg\,kmin}}$ -- Find the subset containing $k$ elements with minimal scores;
  \DashedLine 
  \State $\mathcal{D}^* \leftarrow $random select $k$ samples from $\mathcal{D}$
  \For{$i=1$ to $n$}
	\If {($t \% s$ == $0$) and (not $\mathcal{D}^*$ == $\mathcal{D}$)}
            \State $\mathcal{D}^* \leftarrow \mathcal{D}^* \cup \mathop{\mathrm{arg\,kmin}}\limits_{x \in \mathcal{D}-\mathcal{D}^*}{\mathcal{L}(M, x)}$
	\EndIf
 
	\State $M \leftarrow train(M,E^*,C)$
   \EndFor
\end{algorithmic}
\end{algorithm}
    \end{minipage}
    \end{figure}
    
\subsection{Latent Diffusion Model for UDF}
\label{sec:diffusion}

Once a compact latent space of $\mathbf{z}$ is learned, we conduct latent diffusion. We first compute $\mathbf{z}_t, t = {1, ..., T}$ by gradually introducing Gaussian noise to the data distribution $q\left(\bm{z}_{0: T}\right)$ with a variance schedule. The diffusion process  $p_{\theta}\left(\bm{z}_{0: T}\right)$ then aims to learn the denoising procedure to fit a target distribution by gradual denoising a Gaussian variable through a fixed Markov Chain of length $T$:
\begin{equation}
\begin{split}
q\left(\bm{z}_{0: T}\right)&=q\left(\bm{z}_{0}\right) \prod_{t=1}^{T} q\left(\bm{z}_{t} | \bm{z}_{t-1}\right),\\
p_{\theta}\left(\bm{z}_{0: T}\right)&=p\left(\bm{z}_{T}\right) \prod_{t=1}^{T} p_{\theta}\left(\bm{z}_{t-1} | \bm{z}_{t}\right),
\end{split}
\end{equation}
where $q\left(\bm{z}_{t} | \bm{z}_{t-1}\right)$ and $p_{\theta}\left(\bm{z}_{t-1} | \bm{z}_{t}\right)$ are Gaussian transition probabilities and $\theta$ denotes learnable parameters.

According to~\cite{sohl2015deep, ho2020denoising}, to fit the data distribution by denoising, we have the loss function as follows:
\begin{equation}
\label{eq:diffusion}
  \mathcal{L}_{diffusion} = \mathbb{E}_{\boldsymbol{z}, t, \boldsymbol{\epsilon}\sim\mathcal{N}(0,1)}\left[\left\|\boldsymbol{\epsilon}-\boldsymbol{\epsilon}_{\theta}\left(\bm{z}_t, t\right)\right\|^{2}\right],
\end{equation}
where $\bm{z}_t=\sqrt{\bar{\alpha}_{t}} \boldsymbol{z}_{0}+\sqrt{1-\bar{\alpha}_{t}} \boldsymbol{\epsilon}$, $\boldsymbol{\epsilon}$ is a noise variable, and $t$ is uniformly sampled from $\{1,..., T\}$. The neural network serves a estimator $\boldsymbol{\epsilon}_{\theta}\left(\bm{z}_t, t\right)$.  The diffusion model thus learns the underlying shape distribution of shape embeddings for efficient downstream shape generation tasks.

At the inference stage, we sample $\boldsymbol{z}_{0}$ by 
gradually denoising a noise sampled from the standard normal distribution $N(0, 1)$, and leverage the trained decoder $\mathcal{D}$ to map the denoised code $\boldsymbol{z}_{0}$ back to a 3D UDF field as shown in Fig.~\ref{fig:pipeline}.

\subsection{Conditional Shape Generation}
\label{sec:cond_gen}
We evaluate Surf-D on four conditional shape generation tasks: category-conditioned shape generation, sketch-conditioned shape generation, single-view conditioned shape reconstruction, and text-guided shape generation.

To turn our model into a more flexible conditional shape generator, we utilize the cross-attention mechanism proposed in a latent diffusion model~\cite{rombach2022high}. Specifically, given a condition $c$ from various modalities, we first introduce a task-specific encoder $\tau$ to projects $c$ to an intermediate representation, then map it to the intermediate layers of the UNet via a cross-attention layer:
\begin{equation}
\begin{split}
\text{Attention}(Q, K, V) = \text{softmax}\left(\frac{QK^T}{\sqrt{d_k}}\right) \cdot V \;,
\end{split}
\end{equation}
where $Q = W_Q^{(i)} \cdot \phi_i(z_t), \quad K = W_K^{(i)} \cdot \tau(c), \quad V = W_V^{(i)} \cdot \tau(c).$ $\phi_i(z_t)$ is the output of an intermediate layer of the UNet, $\tau$ is a task-specific encoder, $W_Q^{(i)}$, $W_K^{(i)}$, and $W_V^{(i)}$ are learnable matrices. Based on the condition, Eq.~\ref{eq:diffusion} can be extended as:

\begin{equation}
  \mathcal{L}_{cond} = \mathbb{E}_{\boldsymbol{z}, t, \boldsymbol{\epsilon}\sim\mathcal{N}(0,1)}\left[\left\|\boldsymbol{\epsilon}-\boldsymbol{\epsilon}_{\theta}\left(\bm{z}_t, t, \tau(c)\right)\right\|^{2}\right] \;.
\label{eq:cond_loss}
\end{equation}
With the cross-attention, our models capture the mapping between the conditioned input and the geometry implicitly represented by the latent code.

\subsection{Training Details}
We trained the Point-based AutoEncoder on Deepfashion3D~\cite{zhu2020deep}, ShapeNet~\cite{shapenet2015} and Pix3D~\cite{sun2018pix3d} datasets. The dimension of latent space is $64$. For the input data, we sampled $N=10000$ points on the surface for feature extraction. And the UDF values are clipped at $0.1$ for efficiency in training. We set $\lambda$ in Eq.~\ref{eq:loss_latent} to $0.1$. During the diffusion stage, we adopt the DDPM sampling strategy with a sampling step $T = 1000$. For the conditioning shape generation, we adopt a frozen CLIP~\cite{radford2021learning} text encoder as text encoder $\tau_{text}$ and employ its image encoder as sketch and image encoder~$\tau_{img}$. More details are in the Appendix.
\section{Experiments}
We conduct extensive qualitative and quantitative experiments to demonstrate the generation capability of Surf-D on five tasks: unconditional generation in Sec.~\ref{sec:exp_unconditional}, category-conditioned shape generation, sketch-conditioned shape generation, single-view 3D reconstruction, and text-guided generation in Sec.~\ref{sec:exp_conditional}. We evaluate different AutoEncoder designs in Sec.~\ref{exp:ablation_autoencoder} and latent space interpolation in Sec.~\ref{exp:latent_interpolation}.

\subsection{Dataset}
We evaluate unconditional, category-conditioned, and sketch-conditioned generation tasks on Deepfashion3D~\cite{zhu2020deep} dataset, single-view 3D reconstruction task on Pix3D\cite{sun2018pix3d}dataset, and text-guided generation task on ShapeNet~\cite{shapenet2015} dataset. For the Deepfashion3D dataset, we randomly split garments into train/test splits in a 4:1 ratio. For the Pix3D dataset, we use the provided train/test splits on the chair category following Cheng et al.~\cite{cheng2023sdfusion}. In the absence of official splits for other categories, we randomly split the dataset into disjoint train/test splits in a 4:1 ratio. For the text-guided generation task, we use the Text2shape dataset~\cite{chen2018text2shape} that provides descriptions for the ‘chair’ and ‘table’ categories in the ShapeNet dataset. For a fair comparison, we follow Cheng et al.~\cite{cheng2023sdfusion} to use the train/test splits provided by Xu et al.~\cite{Xu2019disn}.

\begin{figure}[t]
\centering
\begin{overpic}[width=0.96\linewidth]{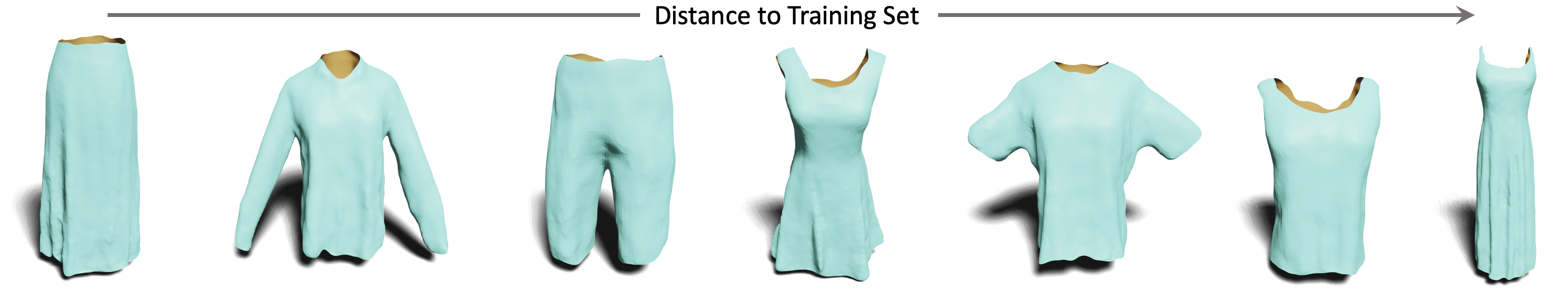}
\end{overpic}
    \caption{\textbf{Samples from unconditional generation.} Our method produces high-quality and diverse shapes. We also calculate their average CD for each object in the training set to confirm that our model is capable of producing unique shapes.}
\label{fig:unconditional}
\end{figure}

\subsection{Unconditional Shape Generation}
\label{sec:exp_unconditional}

\begin{wraptable}[10]{r}{6.3cm}
\caption{\textbf{Quantitative evaluations of unconditional generation on Deepfashion3d.} The units of MMD is $10^{-3}$. COV and 1-NNA are measured in percentages$(\%)$.}
\centering
\resizebox{0.47\textwidth}{!}{
\begin{tabular}{c|c|c|c}
\Xhline{1.2pt}
 & MMD $(\downarrow)$ & COV $(\uparrow)$ & 1-NNA $(\downarrow)$\\
\Xhline{1.2pt}
SDFusion\cite{cheng2023sdfusion}&  14.36&  46.34& 95.73\\
LAS-Diffusion \cite{zheng2023lasdiffusion}& 13.77 & 35.37& 96.34 \\
\Xhline{1.2pt}
\textbf{Ours} & \textbf{8.91} & \textbf{52.44} & \textbf{57.93}\\
\Xhline{1.2pt}
\end{tabular}
}
\label{table:unconditional_quan}
\end{wraptable}

For unconditional generation, we compare our method with SDFusion~\cite{cheng2023sdfusion} and LAS-Diffusion~\cite{zheng2023lasdiffusion}, which recently demonstrated state-of-the-art results on 3D shape generation tasks. We follow Yang et al.~\cite{yang2019pointflow} to use minimum matching distance (MMD), coverage (COV), and 1-nearest neighbor accuracy (1-NNA) based on the Chamfer distance ~\cite{qi2017pointnet} for evaluation. MMD measures quality, COV measures diversity, and 1-NNA measures the similarity of the ground truth and generated distributions. Lower MMD, higher COV, 1-NNA that has a smaller difference to $50\%$ mean better quality. We generate the same number of samples as the test set for evaluation. 

\par As shown in Tab.~\ref{table:unconditional_quan}, our method outperforms baselines in these metrics. Our method gets significantly better scores in the MMD metric which means our method can generate more realistic and detailed shapes. As shown in Fig.~\ref{fig:unconditional}, our method can generate high-quality and diverse clothes. We visualize the generated samples according to the average distance between the generations and each object in the training set.

\begin{figure}[t]
\centering
\begin{overpic}[width=\linewidth]{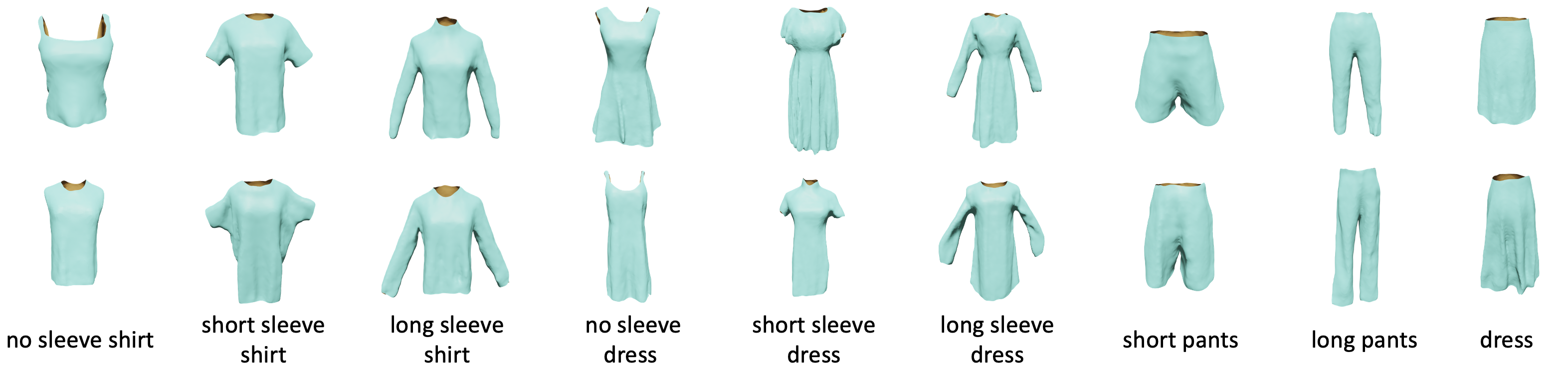}
\end{overpic}
    \caption{\textbf{Qualitative results of Category Conditional Generation.} Our method generates different categories of detailed 3D shapes with high quality and diversity.}
\label{fig:conditional}
\end{figure}

\subsection{Conditional Shape Generation}
\label{sec:exp_conditional}

\paragraph{Category Conditional Generation}
\label{sec:exp_cat_conditional}

\begin{wraptable}[14]{r}{6.4cm}
\caption{\textbf{Quantitative evaluations of category conditional generation on Deepfashion3d.} The units of MMD are $10^{-3}$. COV and 1-NNA are measured in percentages$(\%)$.}
\centering
\resizebox{\linewidth}{!} {
\begin{tabular}{l|l|c|c|c}
\Xhline{1.2pt}
\multirow{2}{*}{Category} & \multirow{2}{*}{Model} & \multirow{2}{*}{MMD $(\downarrow)$} & \multirow{2}{*}{COV $(\uparrow)$} & \multirow{2}{*}{1-NNA $(\downarrow)$} \\
& & & & \\
\Xhline{1.2pt}
\multirow{2}{*}{Shirt} & SDFusion~\cite{cheng2023sdfusion} & 14.67 & 41.67 & 95.83   \\
& \textbf{Ours} & \textbf{8.43} & \textbf{52.78} &  \textbf{59.72}   \\

\Xhline{1.2pt}
\multirow{2}{*}{Dress} & SDFusion~\cite{cheng2023sdfusion} & 12.64 & 47.37 &  84.21   \\
& \textbf{Ours} & \textbf{9.06}  & \textbf{50.00} & \textbf{57.89} \\ 
\Xhline{1.2pt}
\multirow{2}{*}{Pants} &SDFusion~\cite{cheng2023sdfusion} & 17.19 &62.50 & 93.75   \\
& \textbf{Ours} & \textbf{11.27} & \textbf{75.00} & \textbf{62.50}  \\
\Xhline{1.2pt}
\end{tabular}
}
\label{table:category_cond_quan}
\end{wraptable}

In this section, we demonstrate category conditional generation experiments in 9 categories of \textit{no sleeve shirt, short sleeve shirt, long sleeve shirt, no sleeve dress, short sleeve dress, long sleeve dress, dress, short pants, and long pants} objects in Deepfashion3D dataset by conditioning in category information. We use the same evaluation metrics as Tab.~\ref{table:unconditional_quan} to evaluate the generated results. The quantitative comparison of \textit{shirt, dress, pants} objects are in Tab.~\ref{table:category_cond_quan}. As shown in Tab.~\ref{table:category_cond_quan}, our method outperforms SDFusion in all metrics. We also visualize generated 3D shapes in Fig.~\ref{fig:conditional}. Our method generates different categories of detailed 3D shapes with high quality and diversity.

\paragraph{Sketch-guided Shape Generation}
\label{sec:exp_sketch2shape}
\begin{figure}[tb]
\centering
\begin{overpic}[width=\linewidth]{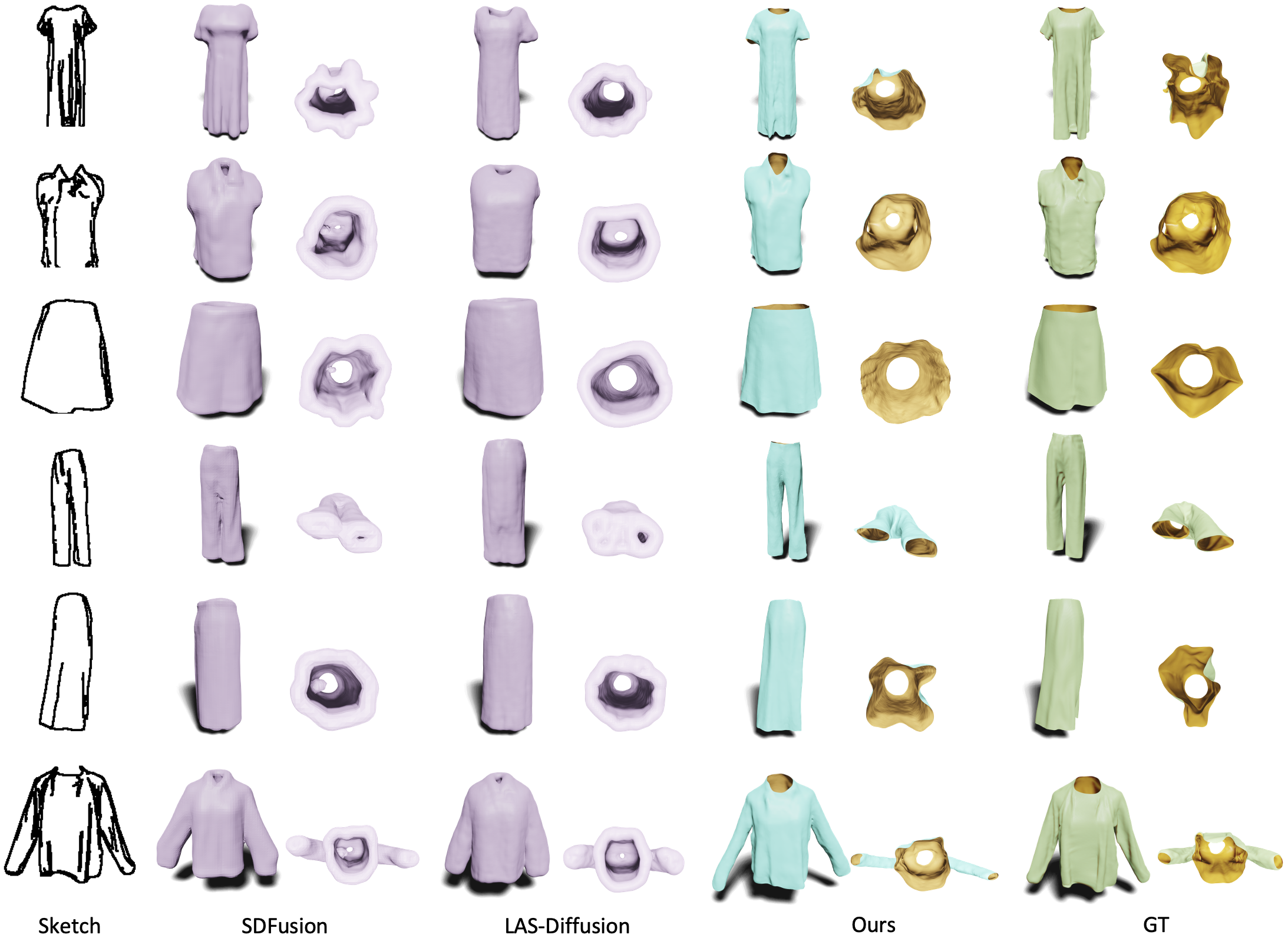}
\end{overpic}
\caption{\textbf{Qualitative results of Sketch Conditioned Generation.} Our method generates higher quality and detailed results aligned with input sketch.}
\label{fig:qual_sketch}
\end{figure}

For sketch-conditioned shape generation, we follow Zheng et al.~\cite{zheng2023lasdiffusion} to generate 2D sketches for each shape in the dataset by rendering its shading images from different views and extracting their edges via the Canny edge detector~\cite{canny1986computational}. The Chamfer distance (CD), Earth mover’s distance (EMD), and the Voxel-IOU are used as evaluation metrics. We compare our method with two state-of-the-art methods for 3D reconstruction and generation, SDFusion~\cite{cheng2023sdfusion} and LAS-Diffusion~\cite{zheng2023lasdiffusion}. Quantitatively, as shown in Tab.~\ref{table:sketch2shape_quan}, our method outperforms other methods on all metrics. As shown in Fig.~\ref{fig:qual_sketch}, SDFusion and LAS-Diffusion generate unrealistic results in open surfaces as the drawback of SDF representations, while our method generates high-quality and detailed results aligned with the input sketch.

\begin{wraptable}[9]{r}{6.4cm}
\caption{\textbf{Quantitative evaluations of sketch conditional generation on Deepfashion3D.} The units of CD, EMD and IOU are $10^{-3}$, $10^{-2}$ and $10^{-2}$.}
\centering
\resizebox{0.5\textwidth}{!}{
\begin{tabular}{c|c|c|c}
\Xhline{1.2pt}
 & CD $(\downarrow)$ & EMD $(\downarrow)$ & IOU $(\uparrow$)\\
\Xhline{1.2pt}
SDFusion\cite{cheng2023sdfusion}&  16.40&  13.33& 34.79\\
LAS-Diffusion \cite{zheng2023lasdiffusion}& 14.37 & 13.02& 34.61 \\
\Xhline{1.2pt}
\textbf{Ours} & \textbf{12.52} & \textbf{11.14} & \textbf{44.06}\\
\Xhline{1.2pt}
\end{tabular}
}
\label{table:sketch2shape_quan}
\end{wraptable}

\begin{figure}[tb]
\centering
\begin{overpic}[width=0.95\linewidth]{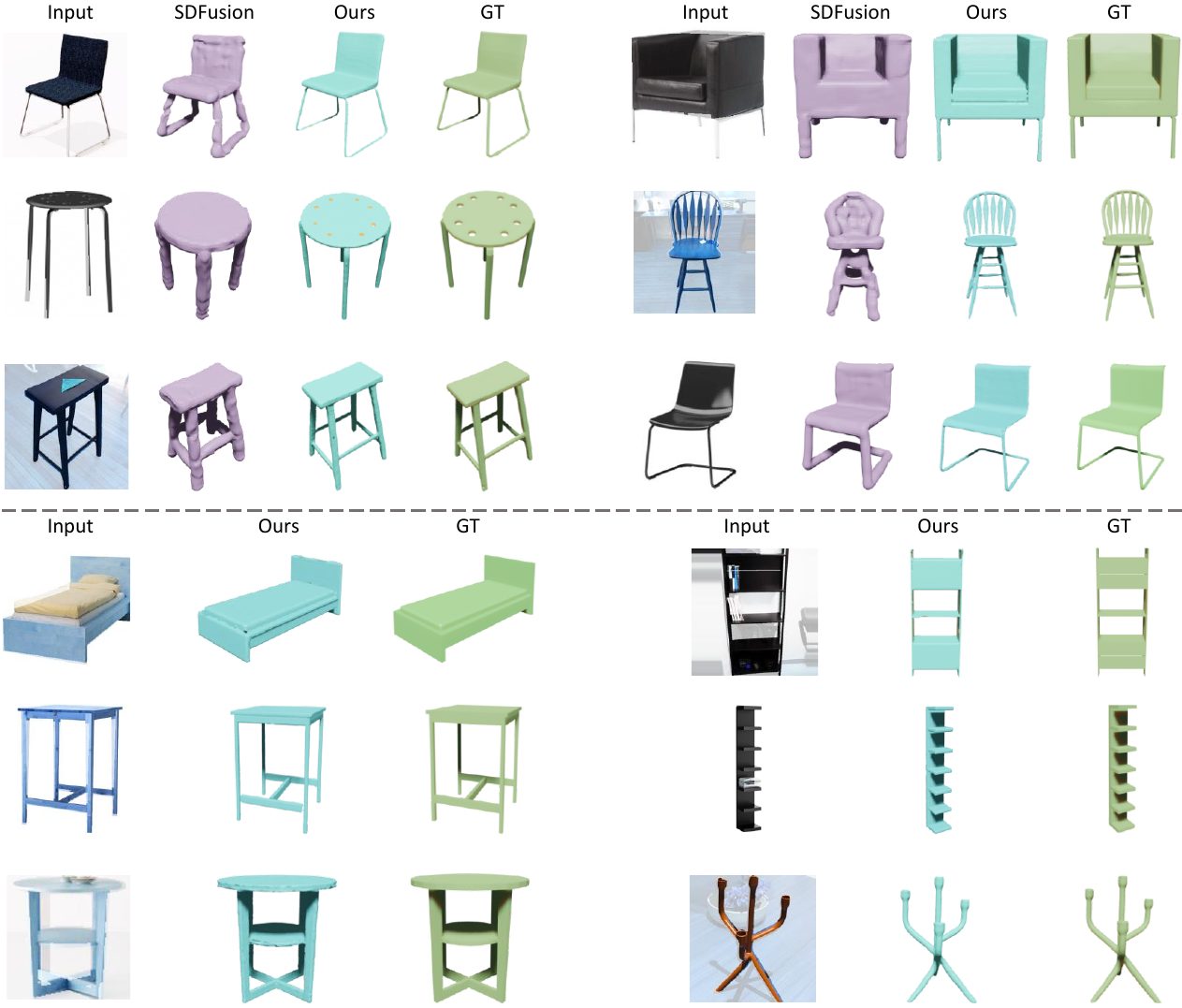}
\end{overpic}
\caption{\textbf{Qualitative results of Single-view Reconstruction.} Surf-D produces high-quality results faithfully aligned with input images.}
\label{fig:img2shape}
\end{figure}

\paragraph{Single-view Shape Reconstruction}
\label{sec:exp_img2shape}
We evaluate 3D shape reconstruction from a single image on the real-world benchmark Pix3D dataset. We compare our method with SDFusion, which is the state-of-the-art method in single-view shape reconstruction in the Pix3D dataset. Since they only release splits in chair objects, we compare with them in chair objects and show our results in other objects. As shown in Fig.~\ref{fig:img2shape}, SDFusion generates over-smooth shapes and cannot handle complex typology shapes, while ours produces high-quality and detailed results even with complex typology shape image inputs.

\paragraph{Text-guided Shape Generation}
\label{sec:exp_text2shape}
\begin{figure}[tb]
\centering
\begin{overpic}[width=0.9\linewidth]{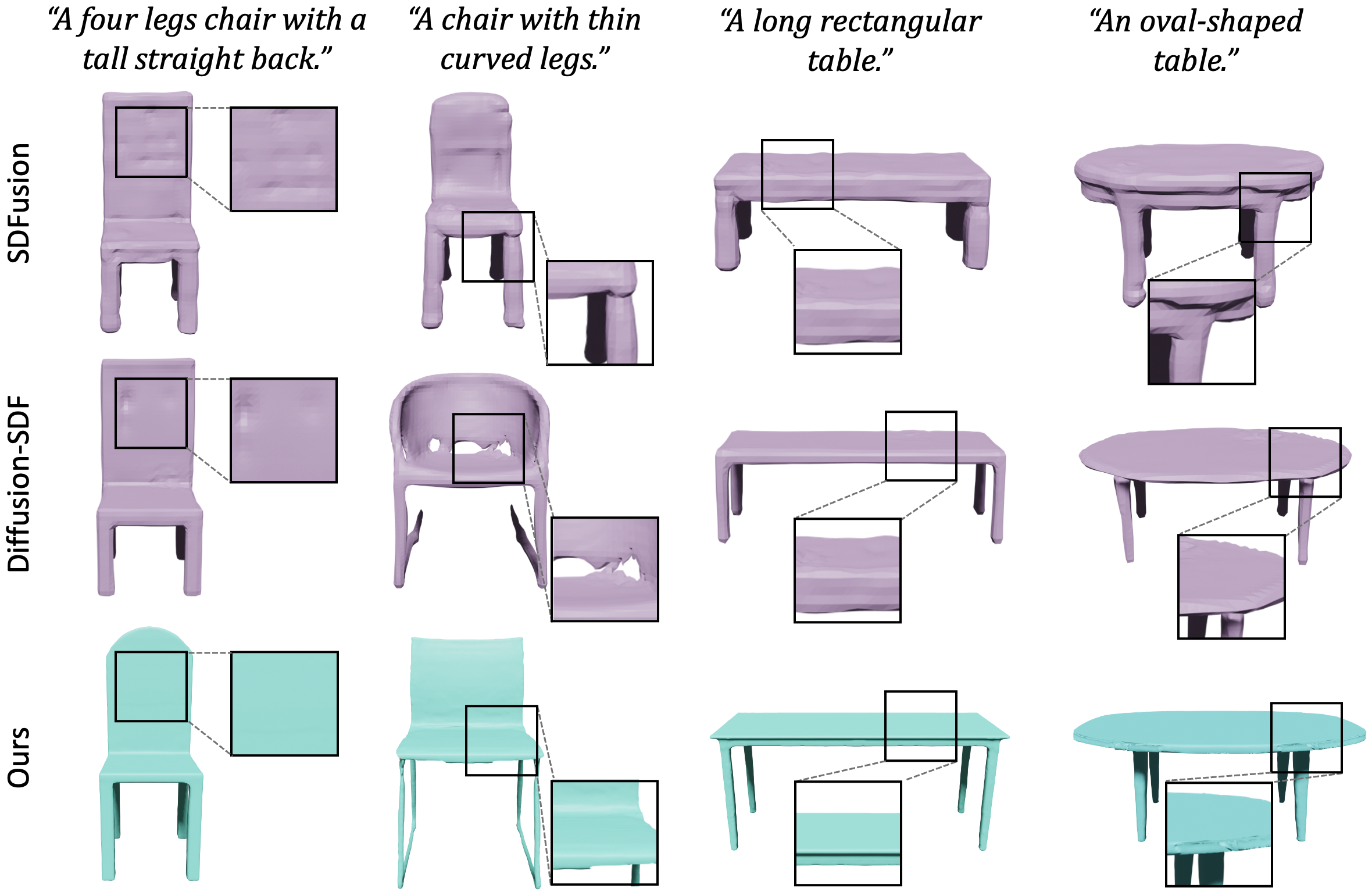}
\end{overpic}
\caption{\textbf{Qualitative comparison of text-to-shape.} Surf-D produces high-quality results aligned with input texts.}
\label{fig:text2shape}
\end{figure}
To evaluate the performance of text-guided shape generation, we conduct experiments on ShapeNet with textual descriptions and compare our method with SOTA methods~SDFusion~\cite{cheng2023sdfusion} and Diffusion-SDF~\cite{li2023diffusion}. As shown in Fig.~\ref{fig:text2shape}, our approach consistently outperforms existing methods by generating shapes that closely align with textual descriptions, which demonstrates its capability to capture various topology and intricate geometry and faithfully translate textual input into high-quality 3D shapes.

\begin{wraptable}[9]{r}{6.4cm}
\caption{\textbf{Memory cost comparison.}}
\centering
\resizebox{0.51\textwidth}{!}{
\begin{tabular}{c|c|c}
\Xhline{1.2pt}
 Methods & Resolution & Model Size \\
\Xhline{1.2pt}
SDFusion~\cite{cheng2023sdfusion}&  $64\times64\times64$ & $4.29$G\\
Diffusion-SDF~\cite{li2023diffusion}& $64\times64\times64$ & $5.86$G \\
LAS-Diffusion~\cite{zheng2023lasdiffusion}& $128\times128\times128$ & $859$M \\
\Xhline{1.2pt}
\textbf{Ours} & \textbf{$\mathbf{512\times512\times512}$} & \textbf{$\mathbf{564}$M} \\
\Xhline{1.2pt}
\end{tabular}
}
\label{table:memory_compare}
\end{wraptable}

\subsection{Grid-based Distance Fields V.S. Continuous Distance Fields}
\label{exp:ablation_autoencoder}
\begin{figure}
\centering
\begin{overpic}[width=0.95\linewidth]{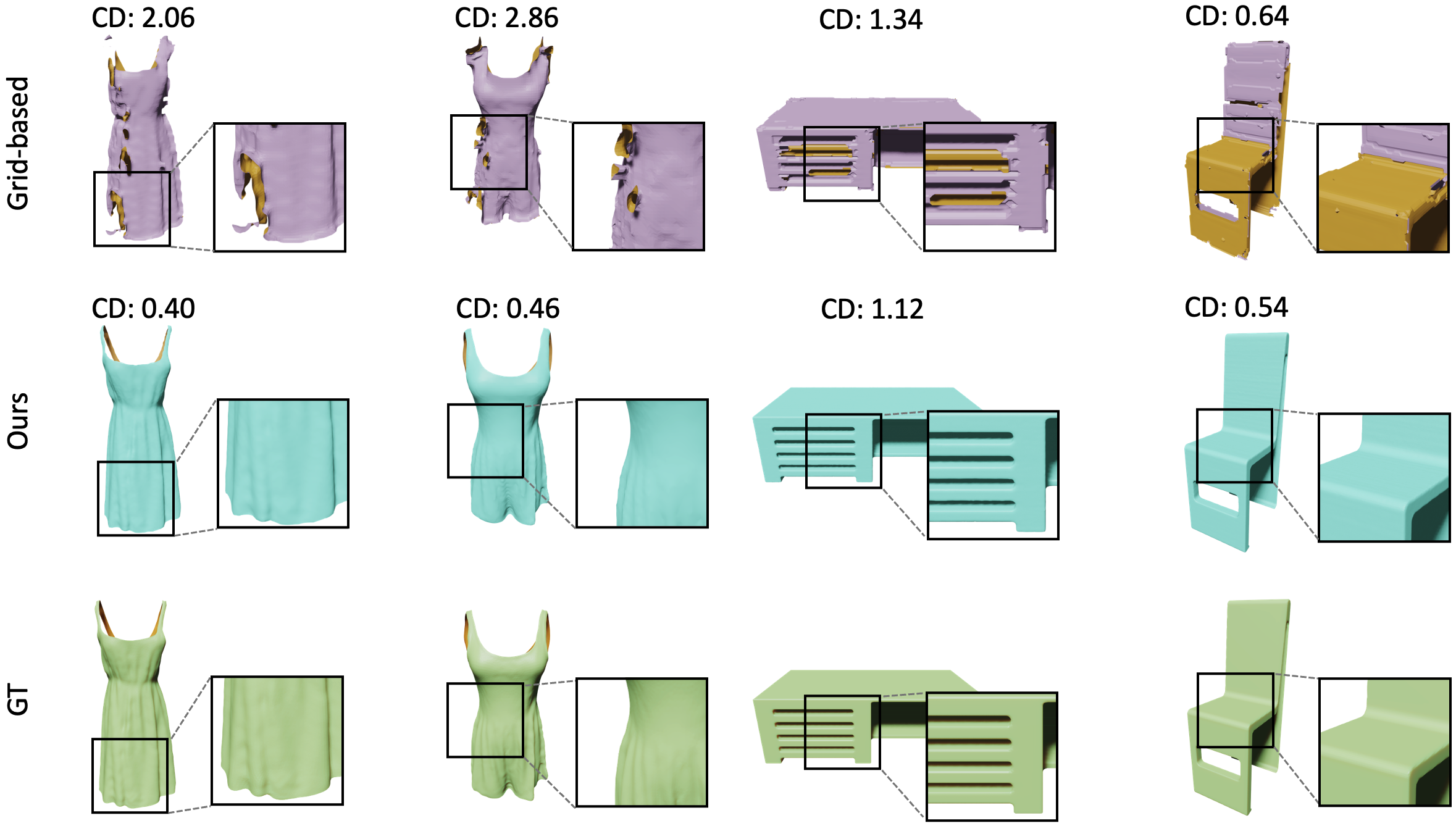}
\end{overpic}
\caption{\textbf{Comparison of different UDF encoding approaches on shape generation.} Compared with the grid-based manner, our method produces high-quality results by learning continuous distance fields.}
\label{fig:autoencoder}
\end{figure}

In this section, we investigate the effectiveness of our point-based AutoEncoder design for shape embedding, in comparison with the grid-based approach. Specifically, we apply the discrete grid-based representation (as used in SDFusion~\cite{cheng2023sdfusion}) to learning UDF, and then compare it with with Surf-D. We test by overfitting the models with corresponding settings to garments and close watertight objects(chairs and tables). As shown in Fig.~\ref{fig:autoencoder}, our model faithfully reconstructs the input shapes with high quality, yielding a lower Chamfer Distance (CD) value, whereas the grid-based model struggles to accurately reconstruct the surface with a higher CD value (the unit of CD is $10^{-3}$). The grid-based representation can not capture implicit fields in detail compared to learning continuous distance fields. A similar observation is also reflected in SDF~\cite{park2019deepsdf}, which is consistent with our observations. Besides, we also compare the resolution and memory cost between our Surf-D with other SOTA methods. As shown in Tab.~\ref{table:memory_compare}, our method achieves higher resolution with lower model size as we learn more compact latent codes compared with other methods. More details of comparison are provided in the Appendix.

\subsection{Latent Interpolation}
\label{exp:latent_interpolation}

\begin{wrapfigure}{r}{5cm}
\centering
\begin{overpic}[width=\linewidth]{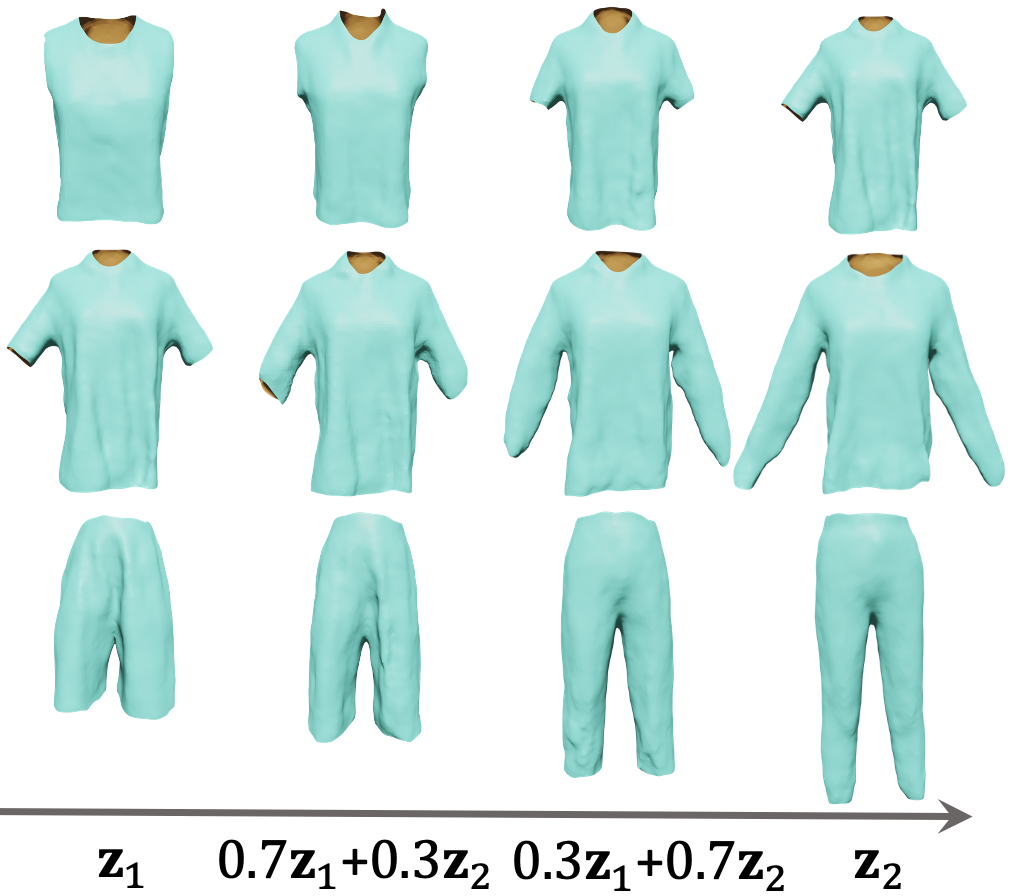}
\end{overpic}
\caption{\textbf{Qualitative results of latent code interpolation.}}
\label{fig:latent}
\end{wrapfigure}

Our point-based auto-encoder provides a compact latent space, which enables the latent diffusion framework to learn the semantics with high efficiency. 
In Fig.~\ref{fig:latent}, we sample latent codes $z_1$ and $z_2$ from latent space and decode the linear interpolated latent codes to show different shapes generated by the naive linear interpolation in the latent space. Even without additional semantic annotation during training, the shape can change smoothly during the latent code interpolation, which demonstrates continuous and high-fidelity shape embedding.

\section{Conclusion}

In this paper, we present Surf-D for generating high-quality 3D shapes as surfaces with arbitrary topologies based on a diffusion model, leveraging Unsigned Distance Fields (UDF) to represent the surfaces as a general shape representation. 
For learning and generating detailed geometry, we propose a latent diffusion model for UDFs by first conducting shape embedding to compress various surfaces into a compact latent space and utilizing diffusion models to capture the shape distribution. With our point-based AutoEncoder that allows gradient querying for each query point in a differentiable manner, our method learns detailed geometry at high resolution. Additionally, the curriculum learning is incorporated for efficient learning of diverse surfaces. Surf-D attains superior quality and controllability in shape generation across various modalities. Extensive experiments validate the effectiveness and efficacy of the proposed method.

\section*{Acknowledgements}
This work is partly supported by the Innovation and Technology Commission of the HKSAR Government under the ITSP-Platform grant (Ref:  ITS/319/21FP) and the InnoHK initiative (TransGP project).

%
%
\bibliographystyle{splncs04}
\bibliography{main}

\clearpage
\appendix

\vspace{1cm}

\begin{center}
\textbf{\LARGE Appendix} 
\end{center}

\begin{figure*}
\vspace{-6mm}
\centering
\begin{overpic}[width=\linewidth]{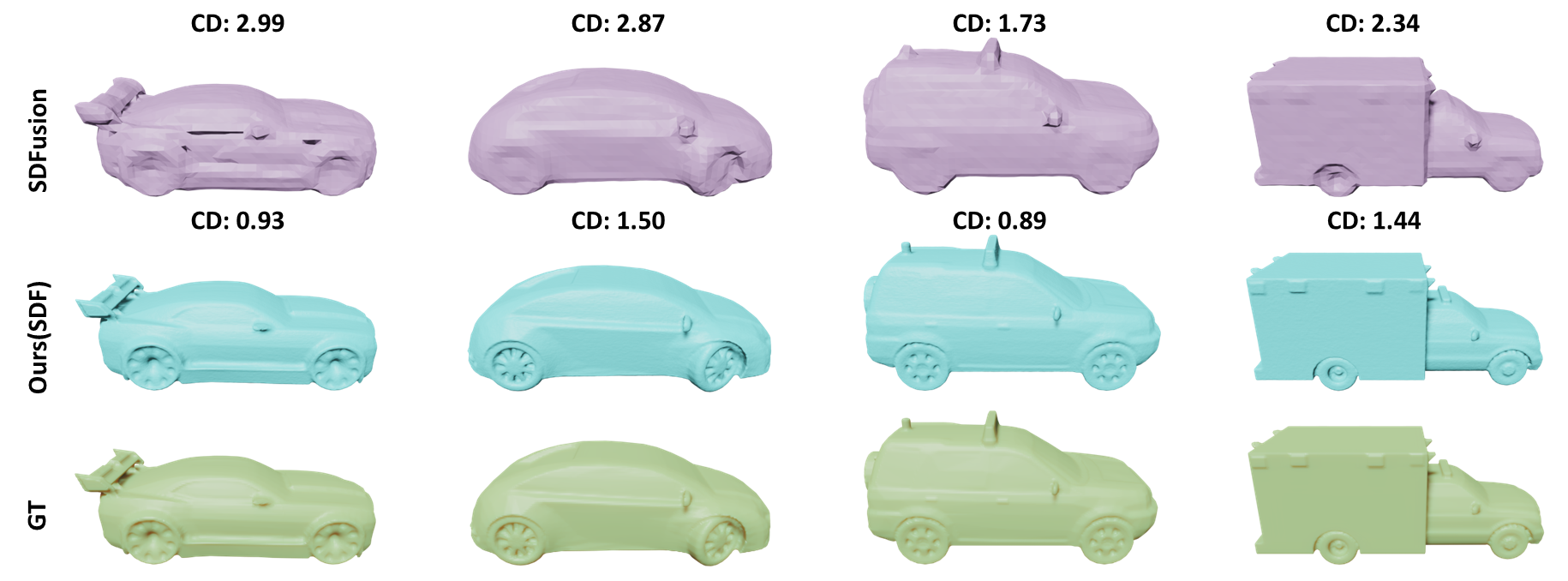}
\end{overpic}
\caption{\textbf{Comparison of learning SDFs.} Our point-based AutoEncoder learns continuous SDFs and produces higher-quality surfaces.}
\label{supp_fig:surfd_on_sdf}
\vspace{-5mm}
\end{figure*}

\begin{figure*}[t]
\vspace{2mm}
\centering
\begin{overpic}[width=\linewidth]{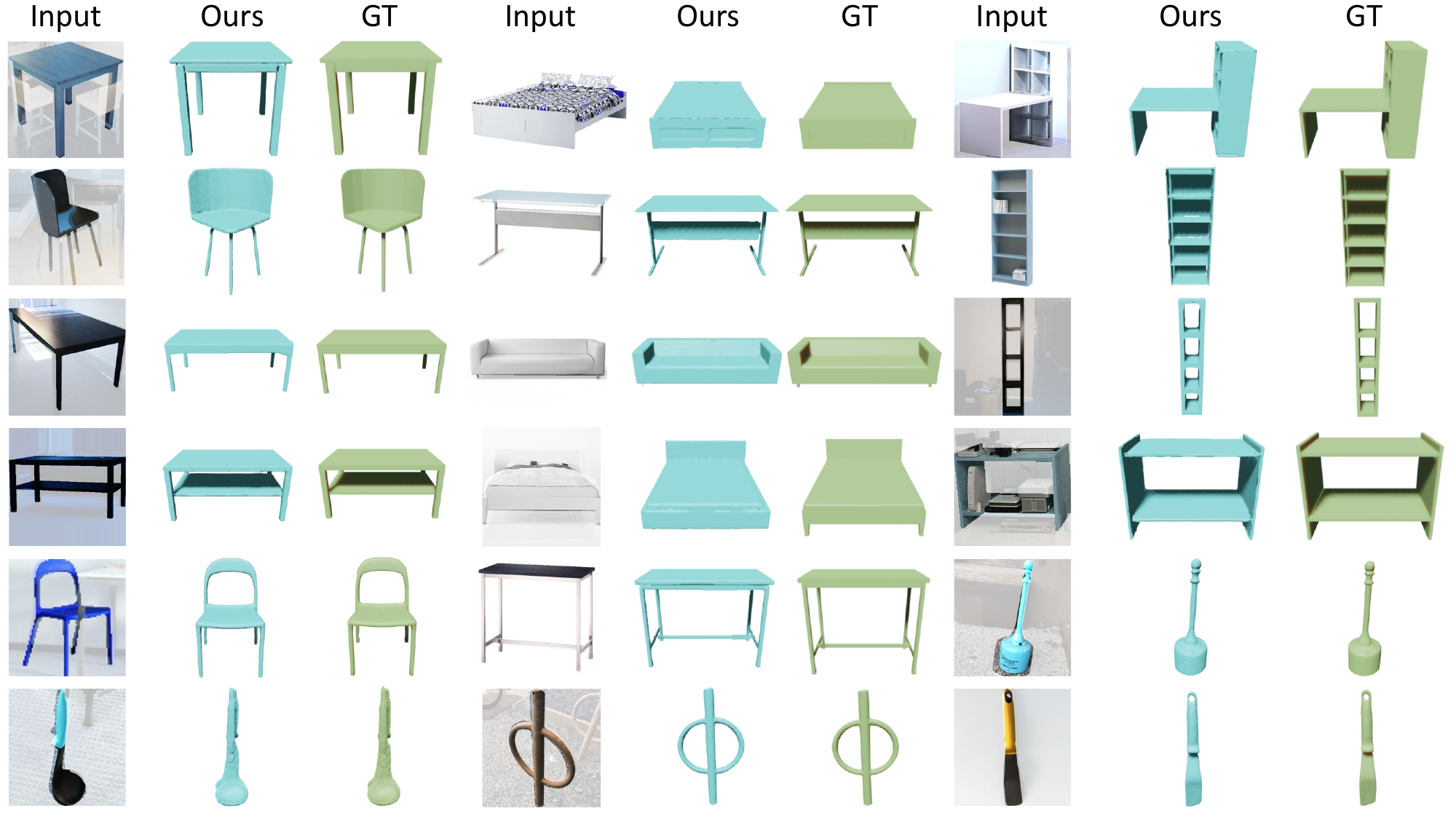}
\end{overpic}
\caption{\textbf{Qualitative results of Single-view Reconstruction.} Surf-D produces high-quality results faithfully aligned with input images.}
\label{supp_fig:pix3d}
\end{figure*}

\begin{figure*}[t]
\centering
\begin{overpic}[width=\linewidth]{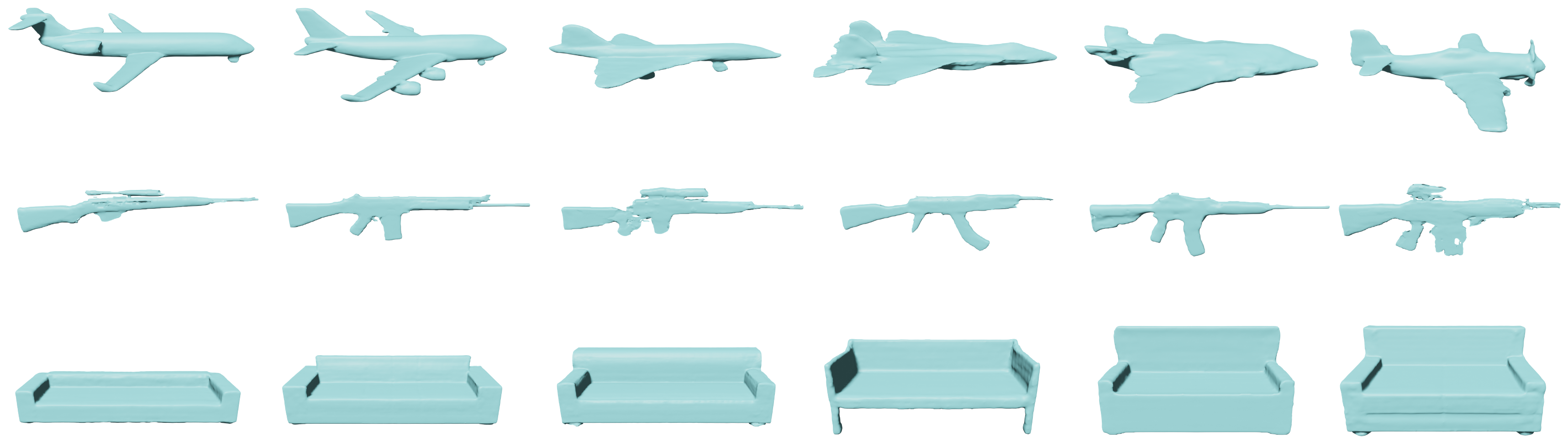}
\end{overpic}
\caption{\textbf{Qualitative results of Unconditional Shape Generation in ShapeNet dataset.} Surf-D produces high-quality, diverse objects with detailed geometry.}
\label{supp_fig:shapenet}
\end{figure*}


This appendix covers: more experimental results~(Sec.~\ref{supp:more_results}); additional ablation studies~(Sec.~\ref{supp:ablation}); implementation details~(Sec.\ref{supp:details}); limitation and future work(Sec.\ref{supp:limitation});). We highly suggest readers visit the including webpage for a more comprehensive review.

\section{More Experimental Results}
\label{supp:more_results}
\subsection{Grid-based Distance Fields V.S. Continuous Distance Fields on SDFs}

In Sec. {\color{red}{4.5}} of the main paper, we demonstrated the effectiveness of our point-based AutoEncoder design for shape embedding of UDFs compared to the grid-based approach~\cite{cheng2023sdfusion}. Here, we further highlight the advantages of continuous distance fields in learning SDFs.

Specifically, we follow SDFusion~\cite{cheng2023sdfusion} to first make the car objects in ShapeNet~\cite{shapenet2015} dataset watertight, and compare our point-based AutoEncoder with grid-based AutoEncoder used in SDFusion for learning SDFs. As shown in Fig.~\ref{supp_fig:surfd_on_sdf}, our pipeline can be adjusted to learn continuous signed distance field and produce higher-quality surfaces than the previous method.

\subsection{More Results on Text-to-Shape Generation}
\begin{wraptable}{r}{5.4cm}
\vspace{-11.5mm}
\caption{\textbf{Quantitative evaluations of single-view reconstruction on Pix3D.} The units of CD, EMD and IOU are $10^{-3}$, $10^{-2}$ and $10^{-2}$.}
\centering
\resizebox{0.45\textwidth}{!}{
\begin{tabular}{c|c|c|c}
\Xhline{1.2pt}
 & CD $(\downarrow)$ & EMD $(\downarrow)$ & IOU $(\uparrow$)\\
\Xhline{1.2pt}
SDFusion\cite{cheng2023sdfusion}&  5.70&  9.45& 52.11\\
\Xhline{1.2pt}
\textbf{Ours} & \textbf{3.72} & \textbf{6.55} & \textbf{66.06}\\
\Xhline{1.2pt}
\end{tabular}
}
\vspace{-8mm}
\label{table:single_view_quan}
\end{wraptable}
We present more quantitative and qualitative results of \mbox{Surf-D} on Pix3D~\cite{sun2018pix3d} and ShapeNet~\cite{shapenet2015} datasets. As shown in Tab.~\ref{table:single_view_quan}, Surf-D outperforms SDFusion on all metrics where our method effectively generates higher quality 3D shapes with arbitrary topology and more geometric details. Qualitatively, as shown in Fig.~\ref{supp_fig:pix3d}, Surf-D consistently yields superior shape reconstruction results that closely correspond to the monocular input images.

\begin{wrapfigure}{r}{5.3cm}
\vspace{-5mm}
\centering
\begin{overpic}[width=\linewidth]{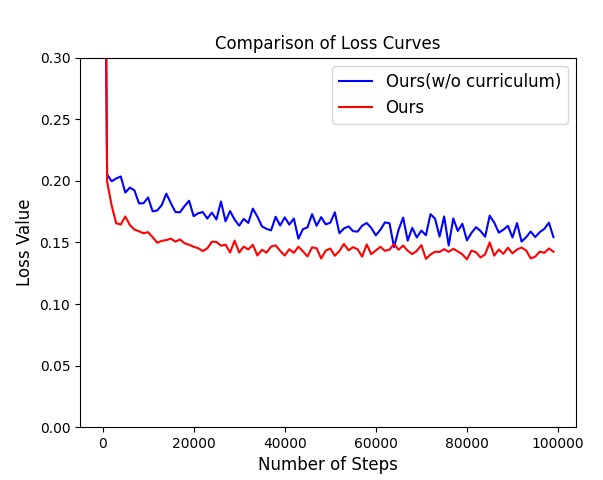}
\end{overpic}
\vspace{-8mm}
\caption{\textbf{Loss Curve Comparison.}}
\label{supp_fig:loss_curve}
\vspace{-8mm}
\end{wrapfigure}

Additionally, we quantitatively compare with other state-of-the-art methods in the text-guided shape generation task. We follow Diffusion-SDF~\cite{li2023diffusion} to use IOU and Total Mutual Difference (TMD) as evaluation metrics. IOU measures the similarity between the generated results and ground truth. TMD measures the diversity of generated results. As shown in Tab.~\ref{table:text2shape}, Surf-D significantly outperforms other methods in terms of the IOU metric, \textit{i.e.}, Surf-D produces more accurate results that faithfully align with the input textual description. Meanwhile, our method achieves competitive performance in TMD.

\subsection{More Results of Unconditional Shape Generation}
\begin{wraptable}{r}{5.4cm}
\vspace{-12mm}
\caption{\textbf{Quantitative evaluations of text-guided shape generation on ShapeNet.}}
\centering
\resizebox{0.45\textwidth}{!}{
\begin{tabular}{c|c|c}
\Xhline{1.2pt}
 & IOU $(\uparrow)$ & TMD $(\downarrow)$ \\
\Xhline{1.2pt}
Liu et al.~\cite{liu2022towards}&  0.160&  0.891\\
AutoSDF~\cite{autosdf2022}&  0.187&  0.581\\
Diffusion-SDF~\cite{li2023diffusion}&  0.194&  \textbf{0.169}\\
\Xhline{1.2pt}
\textbf{Ours} & \textbf{0.282} & 0.175 \\
\Xhline{1.2pt}
\end{tabular}
}
\vspace{-8mm}
\label{table:text2shape}
\end{wraptable}

To further evaluate the generation capability of our method, we randomly select three more category objects(\textit{airplane, rifle, and sofa}) in the ShapeNet\cite{shapenet2015} dataset to train our model separately. The unconditional generation results of these three categories are presented in Fig.~\ref{supp_fig:shapenet}, in which our approach produces high-quality, diverse objects with detailed geometry.

\subsection{Shape for Simulation}
\begin{figure}
\centering
\begin{overpic}[width=0.9\linewidth]{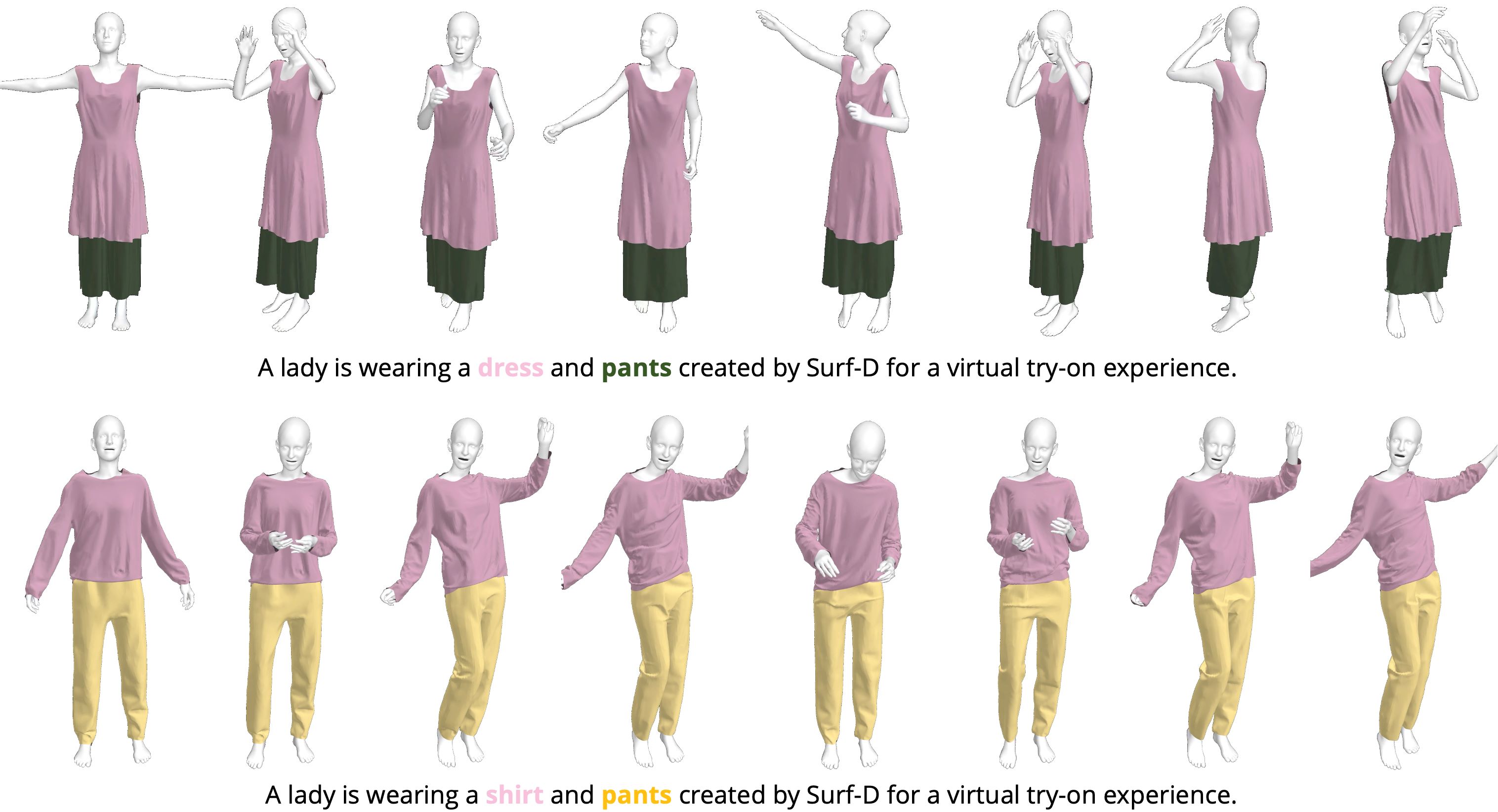}
\end{overpic}
\caption{\textbf{Garment generation for virtual try-on
.} The generated garments can seamlessly integrate into a simulation pipeline for virtual try-on.}
\label{fig:sim}
\end{figure}

Furthermore, we investigate the generated shape for simulation~\cite{liao2024senc, patel2020tailornet}. We show that the generated clothing can be easily integrated into a simulation platform\footnote{\url{https://www.marvelousdesigner.com/}} for virtual try-on, which is shown in Fig.~\ref{fig:sim}. We use the motion sampled from AMASS~\cite{mahmood2019amass} mocap dataset. Please refer to the video on our webpage for the visualization. These compelling results validate our high-quality 3D asset generation.

\section{Additional Ablation Study}
\label{supp:ablation}
\subsection{Ablation of Curriculum Learning Strategy}
\label{supp:curriculum_learning}

We investigate the influence of the proposed curriculum learning strategy during the training of Point-based AutoEncoder. As Fig.~\ref{supp_fig:loss_curve} shows, our model gains a significant improvement when adopting the proposed curriculum learning strategy. Specifically, it contributes to a faster convergence rate and higher reconstruction accuracy during network training. Additionally, since the network is trained in an easy-to-hard setting, our curriculum learning strategy enhances training stability.

In Fig.~\ref{supp_fig:results_cls}, we present the qualitative results of an ablation study on curriculum learning strategy under the same training setting, where the learning strategy promotes a higher-quality geometry.

\subsection{Ablation of the Training Loss Function}
\label{supp:loss}
The choice of $\mathcal{L}_{recon}$ formulation is crucial to achieve better shape embedding results.
In the following, we test three different losses during the learning of distance field of the implicit shapes: (\romannumeral1)~\textit{$L_1$~loss}:~$\mathbb{E} \left[ \|y_{i} -\widetilde{y}_{i} \|_1 \right]$;(\romannumeral2)~\textit{$L_2$~loss}:~$\mathbb{E} \left[ \|y_{i} -\widetilde{y}_{i} \|_2 \right]$; (\romannumeral3)~\textit{our~loss}:~$\mathbb{E} \left[ -{(\bar{y}_{i}\log(\widetilde{y}_{i}) + (1 - \bar{y}_{i})\log(1 - \widetilde{y}_{i}))} \right]$.

\begin{figure}
  \centering
  \vspace{0mm}
  \begin{minipage}[b]{0.48\textwidth}
  \begin{overpic}[width=0.9\linewidth]{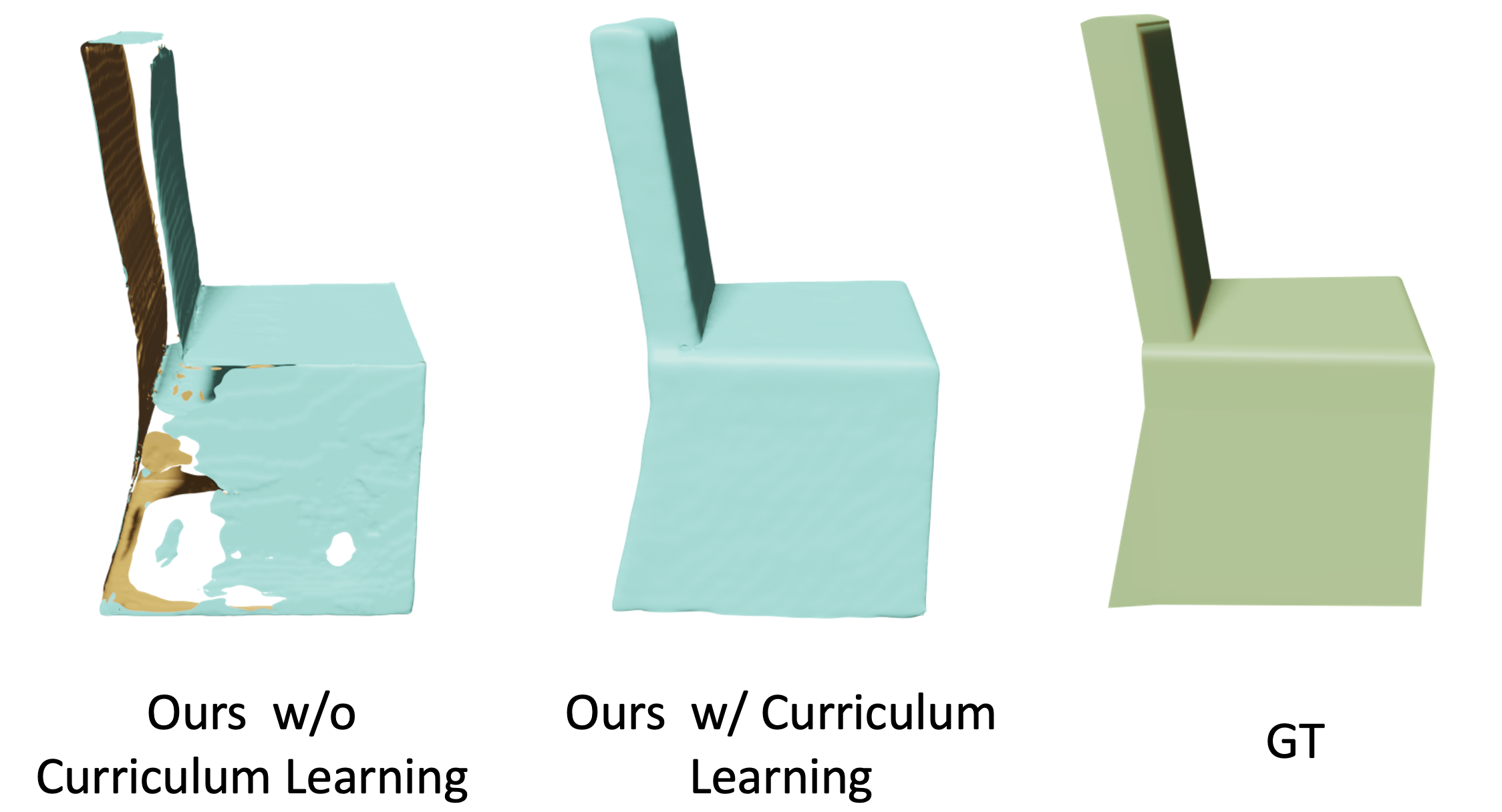}
\end{overpic}
\vspace{-1mm}
\caption{\textbf{Ablation results of curriculum learning strategy.}}
\label{supp_fig:results_cls}
  \end{minipage}
  \hfill 
  \begin{minipage}[b]{0.48\textwidth}
    \begin{overpic}[width=\linewidth]{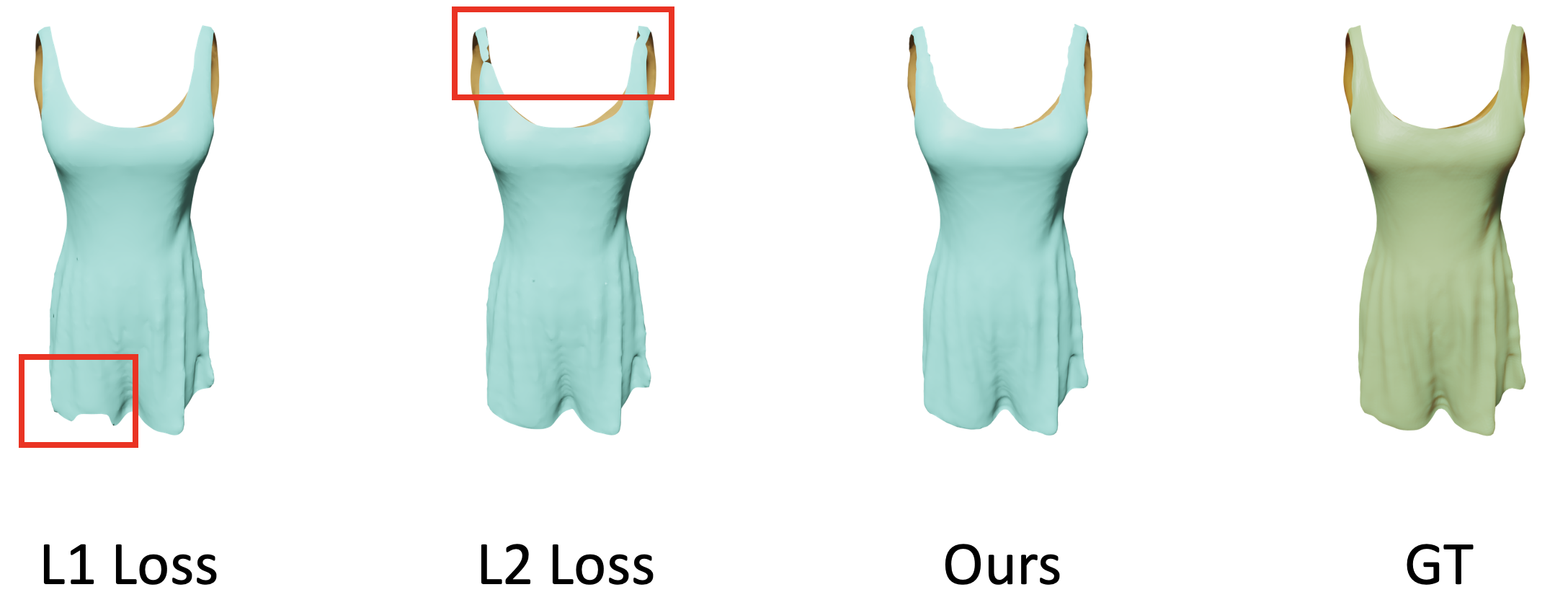}
\end{overpic}
\vspace{-4mm}
\caption{\textbf{Comparison of different reconstruction loss.}}
\label{supp_fig:different_udf_loss}
  \end{minipage}
  \vspace{-6mm}
\end{figure}

As shown in Fig.~\ref{supp_fig:different_udf_loss}, L1 loss tends to lose the edge area in the distance field and results in an incomplete shape. At the same time, L2 loss fails to capture the thin structures and leads to a discontinuity shape. Compared with the two losses, $\mathcal{L}_{recon}$, which is adopted by our method, 
enable the model to effectively capture the continuous distance field and generate high-fidelity and detailed shapes.

\section{Implementation Details}
\label{supp:details}
We provide more implementation details of the query points sampling strategy and the settings of hyperparameters.
\paragraph{Query points sampling strategy for UDF decoder} 
The query points sampling strategy used for the decoder
strongly impacts training effectiveness~\cite{mildenhall2020nerf}. In our sampling strategy, points closer to the surface have a higher probability of being sampled. Specifically, we directly sample $30\%$ of the points on the surface, sample $30\%$ of the points by adding Gaussian noise with $\epsilon$ variance to surface points, obtain $30\%$ with Gaussian noise with $3\epsilon$ variance, and gather the remaining points by uniformly sampling the bounding box in which the object is contained. Since all objects are normalized respectively into the $\left[-1, 1\right]^3$ cube, we consistently set $\epsilon = 0.003$ in our experiments. 
\paragraph{Hyperparameters settings} During the training of the Point-based AutoEncoder, we set the learning rate to $10^{-4}$, and the sample number $k$ in the curriculum learning strategy to $100$. When training the latent diffusion model, we set the learning rate to $10^{-6}$. We use the Adam optimizer~\cite{kingma2014adam} with $\beta_1 = 0.9$ and $\beta_2 = 0.999$. For mesh extraction, we employ MeshUDF~\cite{guillard2022meshudf} with $512 \times 512 \times 512$ resolution to extract meshes from our reconstructed 3D UDF fields.

\paragraph{Dataset size and training time} \textit{Deepfashion3D}~\cite{zhu2020deep} contains $563$ diverse cloth items of $9$ categories. \textit{Pix3D}~\cite{sun2018pix3d} contains $395$ shapes of $9$ categories with $10069$ corresponding images. \textit{Text-to-shape} contains $75,344$ text descriptions for around $10$k chair and table objects. For the training, we use a single 40G A100 GPU. The training time differs on datasets. For \textit{Deepfashion3D}~\cite{zhu2020deep} and \textit{Pix3D}~\cite{sun2018pix3d} dataset, it takes $3$ days to train the AutoEncoder and $2$ days to train the diffusion model. For \textit{Text-to-shape} in\textit{ ShapeNet}~\cite{shapenet2015} dataset, it takes $5$ days to train the AutoEncoder and $3$ days to train the diffusion model.

\section{Limitations and Future Works} 
\label{supp:limitation}
\begin{wrapfigure}{r}{5cm}
\centering
\vspace{-8mm}
\begin{overpic}[width=\linewidth]{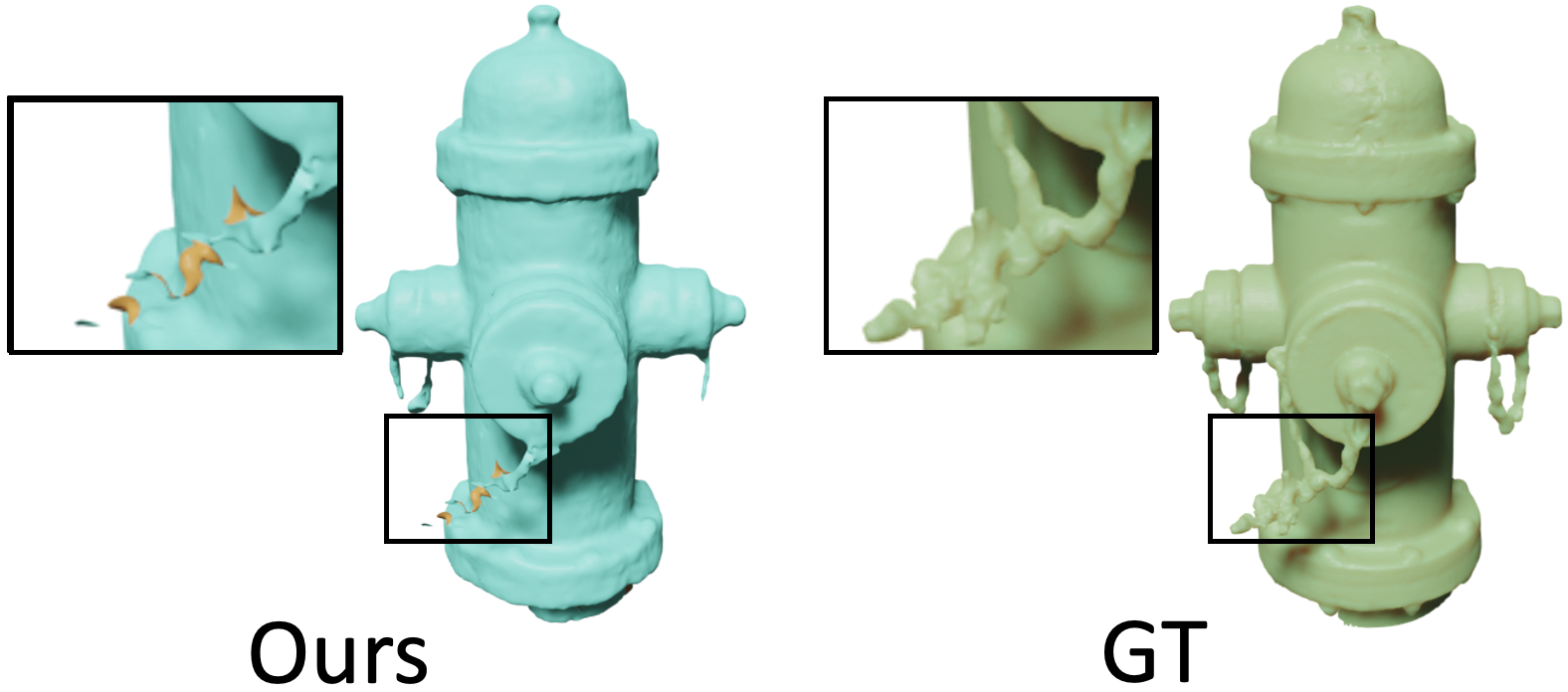}
\end{overpic}
\caption{\textbf{Failure case of Surf-D.}}
\label{supp_fig:failures}
\vspace{-10mm}
\end{wrapfigure}
Although Surf-D demonstrates superior performance in shape generation across multiple modalities, it has some limitations. As shown in Fig.~\ref{supp_fig:failures}, Surf-D is unable to accurately reconstruct the iron chain (see the zoom-in area) because the frequency of geometric details in the iron chain portion is higher than the resolution of the grid for the marching cube, resulting in the loss of geometric details in that part. It would be valuable to explore how to extract mesh from a UDF field efficiently and faithfully. As current mesh extraction methods from the UDF field still require gradients, extracting mesh from the UDF field in a gradient-free way will be an interesting direction. Furthermore, our model requires a relatively large amount of expensive 3D data for capturing the distribution of versatile shapes. Thus, learning shape knowledge from 2D images would significantly alleviate the need for 3D data, which could also be an intriguing research direction.

\end{document}